\documentclass[a4paper,12pt]{article}

\newtheorem{problem}{Problem}
\def\url#1{{\tt #1}}

\global\long\def\A{\mathtt{A}}
\global\long\def\B{\mathtt{B}}
\global\long\def\X{\mathtt{X}}
\global\long\def\Z{\mathtt{Z}}
\global\long\def\ba{\boldsymbol{\alpha}}

\global\long\def\tR{\mathtt{R}}
\global\long\def\bR#1{\mathbb{R}^{#1}}

\global\long\def\tz{\mathbf{t}_{\mathtt{Z}}}

\global\long\def\Rx{\mathtt{R}_{\mathtt{X}}}
\global\long\def\tx{\mathbf{t}_{\mathtt{X}}}

\global\long\def\tai#1{\mathbf{t}_{\mathtt{A}_{#1}}}

\global\long\def\tbi#1{\mathbf{t}_{\mathtt{B}_{#1}}}

\global\long\def\vnorm#1{\left\Vert #1\right\Vert }

\global\long\def\trace{\mathrm{tr}}

\usepackage{amsmath}
\usepackage{amssymb}
\usepackage{amsfonts}
\usepackage{graphicx}
\usepackage{subfig}
\usepackage{rotating}
\usepackage{nicefrac}

\title{\bf Hand-Eye and Robot-World Calibration by Global Polynomial Optimization}

\begin{document}

\author{Jan Heller$^{1}$, Didier Henrion$^{2,3,1}$, Tom\'a\v s Pajdla$^{1}$}

\footnotetext[1]{Czech Technical University, Faculty of Electrical Engineering,
        Karlovo n\'{a}m\v{e}st\'{i} 13, Prague, Czech Republic.}
\footnotetext[2]{CNRS; LAAS; 7 avenue du colonel Roche, F-31077 Toulouse; France.}
\footnotetext[3]{Universit\'e de Toulouse; UPS, INSA, INP, ISAE; UT1, UTM, LAAS; F-31077 Toulouse; France.}

\date{Draft of \today}

\maketitle

\begin{abstract}

The need to relate measurements made by a camera to a different known
coordinate system arises in many engineering applications. Historically,
it appeared for the first time in the connection with cameras mounted
on robotic systems. This problem is commonly known as hand-eye
calibration. In this paper, we present several formulations of hand-eye
calibration that lead to multivariate polynomial optimization problems.
We show that the method of convex linear matrix inequality (LMI) relaxations
can be used to effectively solve these problems and to obtain globally
optimal solutions. Further, we show that the same approach can be
used for the simultaneous hand-eye and robot-world calibration. Finally,
we validate the proposed solutions using both synthetic and real datasets.

\end{abstract}

\section{Introduction}

\begin{figure}[t]
\centering{}\includegraphics[scale=1.2]{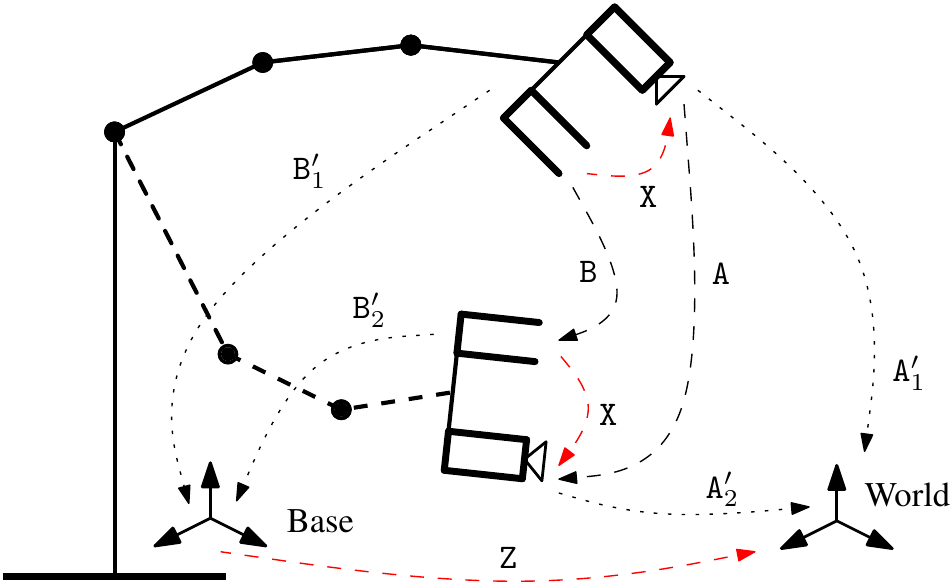}\caption{Hand-eye ($\X$) and robot-world ($\Z$) transformations. A hand-eye
system is depicted in two distinct poses.\label{fig:hand-eye-calibration-schematics}}
\end{figure}

Let us suppose a camera has been mounted on the end-effector---the
hand---of a robotic manipulator. Further, let's suppose that this
hand-eye robotic system has been manipulated into two distinct poses,
see Figure~\ref{fig:hand-eye-calibration-schematics}. Let's denote
the transformation from the camera coordinate system in the first
pose of the rig to the world coordinate system as $\mathtt{A}'_{1}$
and the transformation from the second pose as $\mathtt{A}'_{2}$.
Now, we can express the camera's relative movement from the first
pose to the second one as\addtolength{\arraycolsep}{-0.8mm}
\[
\mathtt{A}=\left(\begin{array}{cc}
\mathtt{R}_{\mathtt{A}} & \mathbf{t}_{\mathtt{A}}\\
\mathbf{0}^{\top} & 1
\end{array}\right)=\left(\begin{array}{cc}
\mathtt{R}_{\mathtt{A}'_{2}} & \mathbf{t}_{\mathtt{A}'_{2}}\\
\mathbf{0}^{\top} & 1
\end{array}\right)^{-1}\left(\begin{array}{cc}
\mathtt{R}_{\mathtt{A}'_{1}} & \mathbf{t}_{\mathtt{A}'_{1}}\\
\mathbf{0}^{\top} & 1
\end{array}\right)=\mathtt{A}'{}_{2}^{-1}\mathtt{A'}_{1},
\]
where $\mathtt{R}_{\mathtt{A}}\in SO(3)$ is a $3\times3$ rotation
matrix and $\mathbf{t}_{\mathtt{A}}\in\bR 3$ is a translation vector.
Analogically, relative movement of the robotic end-effector can be
described as
\[
\mathtt{B}=\left(\begin{array}{cc}
\mathtt{R}_{\mathtt{B}} & \mathbf{t}_{\mathtt{B}}\\
\mathbf{0}^{\top} & 1
\end{array}\right)=\left(\begin{array}{cc}
\mathtt{R}_{\mathtt{B}'_{2}} & \mathbf{t}_{\mathtt{B}'_{2}}\\
\mathbf{0}^{\top} & 1
\end{array}\right)^{-1}\left(\begin{array}{cc}
\mathtt{R}_{\mathtt{B}'_{1}} & \mathbf{t}_{\mathtt{B}'_{1}}\\
\mathbf{0}^{\top} & 1
\end{array}\right)=\mathtt{B'}_{2}^{-1}\mathtt{B}'_{1},
\]
\addtolength{\arraycolsep}{0.8mm}with $\mathtt{B}'_{1},\mathtt{B}'_{2}\in\bR{4\times4}$
being the respective transformations from the end-effector's coordinate
system to the robot base coordinate system. Assuming we know the transformations
$\mathtt{A}$ and $\mathtt{B}$, the problem of finding the rigid
transformation $\X\in\bR{4\times4}$ from the coordinate system of
end-effector to the coordinate system connected with the camera can
be expressed analytically using the following kinematic loop: 
\begin{equation}
\mathtt{AX}=\mathtt{XB}.\label{eq:hand-eye-loop}
\end{equation}
All of the earliest researchers investigating this problem realized
that System~\ref{eq:hand-eye-loop} is underdetermined and that two
poses are not enough to uniquely determine the transformation $\mathtt{X}$.
In \cite{Shiu89}, Shiu and Ahmad showed that at least two relative
motions with non-parallel rotational axes are needed. In practice,
several relative motions are executed, leading to a set of matrices
$\mathtt{A}_{i},\mathtt{B}_{i}$, $i=1,\dots,n$ and to an overdetermined
and---unless we can measure $\mathtt{A}_{i},\mathtt{B}_{i}$ with
perfect accuracy---noisy system of equations
\begin{equation}
\mathtt{A}_{i}\mathtt{X}=\mathtt{X}\mathtt{B}_{i},\quad i=1,\dots,n.\label{eq:hand-eye}
\end{equation}
System~\ref{eq:hand-eye} can be also expressed as\addtolength{\arraycolsep}{-0.8mm}
\begin{equation}
\left(\begin{array}{cc}
\mathtt{R}_{\mathtt{A}_{i}} & \mathbf{t}_{\mathtt{A}_{i}}\\
\mathbf{0}^{\top} & 1
\end{array}\right)\left(\begin{array}{cc}
\mathtt{R}_{\mathtt{X}} & \mathbf{t}_{\mathtt{X}}\\
\mathbf{0}^{\top} & 1
\end{array}\right)=\left(\begin{array}{cc}
\mathtt{R}_{\mathtt{X}} & \mathbf{t}_{\mathtt{X}}\\
\mathbf{0}^{\top} & 1
\end{array}\right)\left(\begin{array}{cc}
\mathtt{R}_{\mathtt{B}_{i}} & \mathbf{t}_{\mathtt{B}_{i}}\\
\mathbf{0}^{\top} & 1
\end{array}\right)\label{eq:hand-eye-detail}
\end{equation}
 \addtolength{\arraycolsep}{0.8mm}and further decomposed into a rotational
matrix equation and translational vector equation
\begin{eqnarray}
\mathtt{R}_{\mathtt{A}_{i}}\mathtt{R}_{\mathtt{X}} & = & \mathtt{R}_{\mathtt{X}}\mathtt{R}_{\mathtt{B}_{i}},\label{eq:hand-eye-rotation}\\
\mathtt{R}_{\mathtt{A}_{i}}\mathbf{t}_{\mathtt{X}}+\mathbf{t}_{\mathtt{A}_{i}} & = & \mathtt{R}_{\mathtt{X}}\mathbf{t}_{\mathtt{B}_{i}}+\mathbf{t}_{\mathtt{X}}.\label{eq:hand-eye-translation}
\end{eqnarray}
Notice that Equation~\ref{eq:hand-eye-rotation} does not depend
on the unknown translation $\tx$. Once rotation $\Rx$ is known,
Equation~\ref{eq:hand-eye-translation} leads to a system of linear
equations in $\tx$ and the translation can be easily determined using
the tools of linear algebra. This fact has been exploited by all of
the earliest solution strategies.

In \cite{Shiu89}, Shiu and Ahmad proposed the first solution to the
hand-eye calibration problem formulated as Equation \ref{eq:hand-eye}.
They used the angle-axis parametrization of the group of rotations
$SO(3)$. The angle-axis parametrization was also used in \cite{Tsai89,Wang92,Park94}.
Chou and Kamel \cite{Chou91} proposed to parametrize Equation~\ref{eq:hand-eye-rotation}
by quaternions. In \cite{Horaud95}, Horaud and Dornaika also used
quaternions to recover $\Rx$ as a minimizer of the same objective
function as \cite{Park94}.

In~\cite{Chen91}, Chen employed the screw motion theory to investigate
the necessary and sufficient conditions for the solutions of Equation~\ref{eq:hand-eye}.
The screw motion theory is based on the fact that a general homogeneous
transformation can be accomplished by means of a translation along
a unique axis and a rotation about the same axis. This is known as
\emph{Chasles theorem} \cite{Chasles31} and such a description of
a rigid motion is known as a \emph{screw}. Chen concluded that in
the case of noisy inputs the computation of $\Rx$ and $\tx$ should
not be decoupled, because otherwise the generality of the result would
be negatively affected. In \cite{Danilidis96,Daniilidis98}, Daniilidis
and Bayro-Corrochano showed how to parametrize Equation~\ref{eq:hand-eye}
using the algebraic counterparts of screws---dual quaternions---and
how to solve for rotational and translational parts of $\X$ simultaneously.
Another simultaneous solution of Equation~\ref{eq:hand-eye} was
proposed by Andreff \emph{et al.} in \cite{Andreff99} using the Kronecker
product. However, their solution needs to be followed by a orthogonalization
of the rotational part $\Rx$. 

Several researchers also proposed iterative solutions to Equation~\ref{eq:hand-eye}.
Zhuang and Shiu~\cite{Zhuang92} proposed an iterative non-linear
method to minimize function $\sum_{i=1}^{n}\vnorm{\mathtt{A}_{i}\mathtt{X}-\mathtt{X}\mathtt{B}_{i}}^{2}$
to simultaneously estimate the rotational and translational parts
of $\X$. As a part of \cite{Horaud95}, Horaud and Dornaika also
proposed a simultaneous iterative method based of quaternions and
Levenberg-Marquardt non-linear optimization \cite{Marquardt63}. They
observed that the method performed well only after introducing two
\emph{ad hoc} selected scaling factors. Both methods need to be provided
with an initial solution estimates and depending on the accuracy of
the estimates may not converge to the global optima. In~\cite{Zhao11},
Zhao suggested an iterative method that is guaranteed to converge
globally optimally based on Second Order Cone Programming. However,
his iterative method does not enforce the orthogonality of the rotation
matrix and thus suffers from the same rotational error propagation
as does the method \cite{Andreff99}. Strobl and Hirzinger \cite{Strobl06}
suggested a novel metric on $SE(3)$ and an iterative method based
on a parametrization of a stochastic model. However, for their method
to perform optimally, some prior information on data noise characteristics
is needed.

In \cite{Zhuang94}, Zhuang \emph{et al.} extended the hand-eye calibration
problem to also include calibration of the robot-world transformation
$\Z$, see Figure~\ref{fig:hand-eye-calibration-schematics}. Their
method uses quaternion rotation representation to solve an equation
analogical to the Equation~\ref{eq:hand-eye},
\begin{equation}
\A_{i}'\X=\mathtt{ZB}_{i}',\, i=1,\dots,m,\label{eq:robot_world}
\end{equation}
where $\mathtt{Z}\in\bR{4\times4}$ represents the transformation
from the robot base coordinate system to the world coordinate system.
In this case, the kinematic loop is closed using the absolute camera
and robot poses $\A_{i}',\B_{i}'$. Dornaika and Horaud~\cite{Dornaika98}
suggested a different solution, also based on quaternions. In \cite{Li10},
Li \emph{et al.} proposed two different solutions based on Kronecker
product and dual quaternions, analogous to the solutions to the hand-eye
calibration problem in~\cite{Andreff99} and~\cite{Daniilidis98},
respectively.

Recently, several authors \cite{Seo09,Kim10-2,Heller11,Heller12,Ruland12}
proposed hand-eye calibration methods that use image measurements
directly, instead of using them to compute matrices $\A'_{i}$ as
a pre-step. While such an approach can indeed eliminate the errors
resulting from explicit computation of matrices $\A'_{i}$, it is
only applicable in situations where the image correspondences are
available, which may not always be the case.

In this paper, we propose a set of iterative methods to solve hand-eye
and robot-world calibration based on Equations~\ref{eq:hand-eye}
and~\ref{eq:robot_world} that do not require initial estimates and
provide globally optimal solutions in $L_{2}$-norm. Further, these
methods solve for the rotational and translational part simultaneously.
This is achieved by formulating the hand-eye and robot-world calibration
problems as multivariate polynomial optimization problems over semialgebraic
sets and by solving them using the method of convex linear matrix
inequality (LMI) relaxations~\cite{Lasserre01}. Besides providing
global optimizers, one of the main advantages of LMI relaxations method
is the fact that it naturally handles algebraic as well as semialgebraic
constraints that are inherent to the most rotation parametrizations.

First, we review the method of convex LMI relaxations. Next, we formulate
a set of hand-eye and robot-world calibration polynomial optimization
problems. Finally in the experimental section, we provide a few implementation
details and show the performance of the proposed solution using both
synthetic and real data calibration scenarios.

\section{Convex LMI Relaxations}

Let $p_{i}(\mathbf{x})$, $i=0,1,\dots,\ell$ be scalar multivariate
polynomials in $\mathbf{x}=(x_{1},x_{2},\dots,x_{m})^{\top}\in\bR m$,
\emph{i.e.}, $p_{i}(\mathbf{x})\in\mathbb{R}[\mathbf{x}]$. Formally,
the problem of multivariate polynomial optimization can be stated
as follows:\vspace{0.2cm}\begin{problem}(Multivariate polynomial optimization)
\[
\begin{array}{rl}
\textrm{minimize} & p_{0}(\mathbf{x})\\
\textrm{subject to} & p_{i}(\mathbf{x})\geq0,\, i=1,\dots,\ell,\\
\textrm{where} & \mathbf{x}=(x_{1},x_{2},\dots,x_{m})^{\top}\in\bR m,\\
 & p_{0}(\mathbf{x}),p_{i}(\mathbf{x})\in\mathbb{R}[\mathbf{x}].
\end{array}
\]
\label{prob:mpoly_opt}\end{problem}Typically, this is a non-convex
problem with many local minima. In theory, polynomial optimization
problems can be handled using tools of elementary calculus, however,
finding the global minimizer $\mathbf{x}^{*}$ is---in general---an
NP-hard problem \cite{Murty87}. In~\cite{Lasserre01}, Lasserre
cast the problem polynomial optimization over finite-dimensional semialgebraic
set $S=\{\mathbf{x}\in\bR m\,|\, p_{i}(\mathbf{x})\geq0,i=1,\dots,\ell\}$
as the problem of linear optimization over the infinite-dimensional
set of probability measures supported on $S$. In \cite{Putinar93},
Putinar proved that such probability measures can be represented via
sequences $\mathbf{y}=(y_{\alpha})_{\alpha\in\mathbb{N}}$ of its
moments. Using this result, Lasserre showed that by truncating these
sequences one can construct a hierarchy of convex relaxations $\mathcal{P}_{1},\mathcal{P}_{2},\dots$
that produces a monotonically non-decreasing sequence of lower bounds
on Problem~\ref{prob:mpoly_opt} that converge to the global minimum.
He also showed, that the series of the respective global optimizers
$\mathbf{x}_{1}^{*},\mathbf{x}_{2}^{*},\dots$ of problems $\mathcal{P}_{1},\mathcal{P}_{2},\dots$
asymptotically converges to $\mathbf{x}^{*}$, $\lim_{i\rightarrow\infty}\mathbf{x}_{i}^{*}=\mathbf{x}^{*}$,
and that under mild conditions global optimality of a relaxation can
be detected and the global minimizers can be extracted by linear algebra
from the solutions of the relaxation. Practically, $(\mathbf{x}_{i}^{*})_{i\in\mathbb{N}}$
converges to $\mathbf{x}^{*}$ in finitely many steps, \emph{i.e.},
there exists $j\in\mathbb{N}$, such that $\mathbf{x}_{j}^{*}=\mathbf{x}^{*}$.
Problems for which the finite convergence does not occur are in some
sense degenerate and exceptional \cite{Nie12}. The hierarchy $\mathcal{P}_{1},\mathcal{P}_{2},\dots$
is sometimes also called \emph{Lasserre's LMI hierarchy}. 

Now, we will show how to formulate the relaxations $\mathcal{P}_{1},\mathcal{P}_{2},\dots$
as semidefinite programs (SDP) solvable by any convenient SDP solver.
In order to do that, let us define the linearization operator $L_{\mathbf{y}}\colon\mathbb{R}[\mathbf{x}]\rightarrow\mathbb{R}[\mathbf{y}]$
(also called \emph{Riesz functional}) which takes a polynomial $p(\mathbf{x})$
and substitutes a new variable $y_{k_{1}k_{2}\dots k_{m}}\in\mathbb{R}$
for every monomial $x_{1}^{k_{1}}x_{2}^{k_{2}}\dots x_{m}^{k_{m}}$.
First, The LMI relaxation $\mathcal{P}_{\delta}$ of order $\delta$
is built by linearizing all monomials $x_{1}^{k_{1}}x_{2}^{k_{2}}\dots x_{m}^{k_{m}}$
of the objective function $p_{0}$ up to degree $2\delta$, \emph{i.e.},
$k_{1}+k_{2}+\cdots+k_{m}\leq2\delta$. If the objective function
contains monomials of a higher degree, one has to start with a relaxation
of a higher order. Next, let $\mathbf{v}_{\delta}(\mathbf{x})$ be
the vector of all monomials up to degree~$\delta$. The semialgebraic
set $S$ is relaxed by introducing $\ell$ LMI constraints $L_{\mathbf{y}}(p_{i}(\mathbf{x})\mathbf{v}_{\delta-1}(\mathbf{x})\mathbf{v}_{\delta-1}(\mathbf{x})^{\top})\succeq0$%
\footnote{Notation ``$\mathtt{M}\succeq0$'' stands for ``$\mathtt{M}$ is
positive semidefinite matrix''.%
}. Finally, we add the so-called LMI moment matrix constraint $L_{\mathbf{y}}(\mathbf{v}_{\delta}(\mathbf{x})\mathbf{v}_{\delta}(\mathbf{x})^{\top})\succeq0$.
Formally, the LMI relaxation $\mathcal{P}_{\delta}$ of order $\delta$
can be written as

\vspace{0.2cm}\begin{problem}(the LMI relaxation $\mathcal{P}_\delta$  of order $\delta$)\vspace{-0.4cm}

\[
\begin{array}{rl}
\textrm{minimize} & L_{\mathbf{y}}(p_{0}(\mathbf{x}))\\
\textrm{subject to} & L_{\mathbf{y}}(p_{i}(\mathbf{x})\mathbf{v}_{\delta-1}(\mathbf{x})\mathbf{v}_{\delta-1}(\mathbf{x})^{\top}\succeq0,\, i=1,\dots,\ell,\\
 & L_{\mathbf{y}}(\mathbf{v}_{\delta}(\mathbf{x})\mathbf{v}_{\delta}(\mathbf{x})^{\top})\succeq0.
\end{array}
\]

\label{prob:mpoly_sdp}\end{problem}Since there are exactly $d=\tbinom{m+2\delta}{m}$
monomials in $\mathbf{x}\in\mathbb{R}^{m}$ up to degree $2\delta$,
SDP Problem~\ref{prob:mpoly_sdp} will have $\mathbf{y}\in\mathbb{R}^{d}$
linear variables. See~\cite{Lasserre01} for the technical justification
of this procedure.

\section{Hand-Eye Calibration}

In this section, three formulation of the hand-eye calibration problem
are presented. In the first two, we formulate two parametrizations
of of the following minimization problem:
\[
\min_{\X\in SE(3)}\sum_{i=1}^{n}\vnorm{\mathtt{A}_{i}\mathtt{X}-\mathtt{X}\mathtt{B}_{i}}^{2},
\]
\emph{i.e.}, the minimization the Frobenius norm 
\[
\vnorm{\mathtt{M}}=\sqrt{{\textstyle \sum_{i=1}^{m}\sum_{j=1}^{n}|m_{ij}|^{2}}}=\sqrt{\trace(\mathtt{M}^{\top}\mathtt{M})}
\]
on the special Euclidean group 
\[
SE(3)=\left\{ \left.\left(\begin{array}{cc}
\mathtt{R} & \mathbf{t}\\
\mathbf{0}^{\top} & 1
\end{array}\right)\right|\mathtt{R}\in SO(3),\mathbf{t}\in\bR 3\right\} .
\]
The third formulation uses the dual quaternion parametrization to
minimize the vector $L_{2}$-norm
\[
\min_{\tilde{\mathbf{q}}_{\X}}\sum_{i=1}^{n}\vnorm{\hat{\mathbf{a}}_{i}\otimes\hat{\mathbf{q}}_{\X}-\hat{\mathbf{q}}_{\X}\otimes\hat{\mathbf{b}}_{i}}^{2},
\]
where $\hat{\mathbf{a}}_{i},\hat{\mathbf{b}}_{i},\hat{\mathbf{q}}_{\X}$
are the dual quaternion representations of $\mathtt{A}_{i}$,$\mathtt{B}_{i}$,
and $\mathtt{X}$, respectively, and $\otimes$ is the dual quaternion
multiplication.

All three formulations lead to the multivariate polynomial optimization
problems. However, for the sake of brevity we will not cite the explicit
form of the polynomials. The fact that these formulations are indeed
polynomial can be easily checked by any tool for symbolic algebra
computation.

\subsection{Orthonormal parametrization}

As is the case with all linear maps on finite-dimensional vector spaces,
a rotation can be always expressed by a matrix $\tR$, in this case
of size $\mathrm{3}\times\mathrm{3}$. Since a rotation maps orthonormal
basis of $\bR 3$ to another orthonormal basis, the columns $\mathbf{u},\mathbf{v},\mathbf{w}\in\bR 3$
of the matrix $\tR=\left(\mathbf{u},\mathbf{v},\mathbf{w}\right)$
themselves must form an orthonormal basis. The fact that the columns
of $\tR$ form a orthonormal basis can be also written as
\[
\begin{array}{cc}
\mathbf{v}^{\top}\mathbf{v}=1, & \mathbf{u}^{\top}\mathbf{u}=1,\\
\mathbf{v}^{\top}\mathbf{u}=0, & \mathbf{v}\times\mathbf{u}=\mathbf{w}.
\end{array}
\]
This constitutes 6 constraints on the elements of the matrix $\tR$,
leaving it with 3 degrees of freedom. Using these constraints, we
can parametrize the homogeneous transformation $\X$ as
\[
\X(\mathbf{u},\mathbf{v},\mathbf{t})=\left(\begin{array}{cc}
\tR(\mathbf{u},\mathbf{v}) & \mathbf{t}\\
\mathbf{0}^{\top} & 1
\end{array}\right),
\]
where $\tR(\mathbf{u},\mathbf{v})=\left(\mathbf{u},\mathbf{v},\mathbf{u}\times\mathbf{v}\right)$.
This leads to the following parametrization of the hand-eye calibration
problem:

\vspace{0.2cm}\begin{problem}({\it uvhec} method)\vspace{-0.3cm}

\[
\begin{array}{rl}
\textrm{minimize} & f_{1}(\mathbf{u}_{\X},\mathbf{v}_{\X},\tx)=\\
 & \hspace{0.5cm}\sum_{i=1}^{n}\vnorm{\A_{i}\X(\mathbf{u}_{\X},\mathbf{v}_{\X},\tx)-\X(\mathbf{u}_{\X},\mathbf{v}_{\X},\tx)\B_{i}}^{2}\\
\textrm{subject to} & \mathbf{u}_{\X}^{\top}\mathbf{u}_{\X}=1,\mathbf{v}_{\X}^{\top}\mathbf{v}_{\X}=1,\mathbf{u}_{\X}^{\top}\mathbf{v}_{\X}=0.
\end{array}
\]
\label{prob:uvhec}\end{problem}The objective function $f_{1}$ of
Problem~\ref{prob:uvhec} is a polynomial function of degree 4 and
it is composed of 123 monomials in 9 variables.

\subsection{Quaternion parametrization}

Quaternions, $\mathbb{Q}$, form a four-dimensional associative normed
division algebra over the real numbers. A quaternion $\mathbf{q}\in\mathbb{Q}$
consist of a real part and an imaginary part and is usually denoted
as 
\[
\mathbf{q}=q_{1}+q_{2}\mathbf{i}+q_{3}\mathbf{j}+q_{4}\mathbf{k},
\]
where $\mathbf{i},\mathbf{j},\mathbf{k}$ are the imaginary units
such that $\mathbf{i}^{2}=\mathbf{j}^{2}=\mathbf{k}^{2}=\mathbf{i}\mathbf{j}\mathbf{k}=-1.$
As a set, quaternions are equal to $\bR 4$ and it is sometimes useful
to write them as vectors
\[
\mathbf{q}\equiv\left(q_{1},q_{2},q_{3},q_{4}\right)^{\top}=\left(q_{1},\bar{\mathbf{q}}^{\top}\right)^{\top},
\]
where $\bar{\mathbf{q}}\in\bR 3$ is the imaginary part of the quaternion.
Addition of two quaternions $\mathbf{p},\mathbf{q}\in\mathbb{Q}$,
$\mathbf{p}=\left(p_{1},\bar{\mathbf{p}}^{\top}\right)^{\top}$, $\mathbf{q}=\left(q_{1},\bar{\mathbf{q}}^{\top}\right)^{\top}$
is equivalent to addition in $\bR 4$. Quaternion multiplication,
however, does not have a counterpart operation on vector spaces:
\[
\mathbf{p}*\mathbf{q}=(p_{1}q_{1}-\bar{\mathbf{p}}^{\top}\bar{\mathbf{q}},(p_{1}\bar{\mathbf{q}}+q_{1}\bar{\mathbf{p}}+\bar{\mathbf{p}}\times\bar{\mathbf{q}})^{\top})^{\top}.
\]
Because the group of unit quaternions with multiplication, modulo
the multiplication by $-1$, is isomorphic to the group of rotations
with composition, they can be used to represent rotations. Rotation
about axis $\ba=(\alpha_{1},\alpha_{2},\alpha_{3})^{\top}$, $||\ba||=1$,
by angle $\theta$ is represented by $\mathbf{q}\in\mathbb{Q}$ as
\[
\mathbf{q}=\cos\frac{\theta}{2}+(\alpha_{1}\mathbf{i}+\alpha_{2}\mathbf{j}+\alpha_{3}\mathbf{k})\sin\frac{\theta}{2}.
\]
The unity of a quaternion can be expressed using the quaternion conjugate
$\mathbf{q}^{*}=\left(q_{1},-\bar{\mathbf{q}}^{\top}\right)^{\top}$
as $\mathbf{q}^{*}*\mathbf{q}=1$ or using the inner product as $\mathbf{q}^{\top}\mathbf{q}=1$.
Transformation $\X$ can be parametrized using the unit quaternion
$\mathbf{q}=(q_{1},q_{2},q_{3},q_{4})$ as 
\[
\X(\mathbf{q},\mathbf{t})=\left(\begin{array}{cc}
\tR(\mathbf{q}) & \mathbf{t}\\
\mathbf{0}^{\top} & 1
\end{array}\right),
\]
where
\[
\mathtt{R}(\mathbf{q})=\left(\begin{array}{ccc}
{ q_{1}^{2}+q_{2}^{2}-q_{3}^{2}-2q_{4}^{2}} & { 2q_{2}q_{3}-2q_{4}q_{1}} & { 2q_{2}q_{4}+2q_{3}q_{1}}\\
{ 2q_{2}q_{3}+2q_{4}q_{1}} & { q_{1}^{2}-q_{2}^{2}+q_{3}^{2}-q_{4}^{2}} & { 2q_{3}q_{4}-2q_{2}q_{1}}\\
{ 2q_{2}q_{4}-2q_{3}q_{1}} & { 2q_{3}q_{4}+2q_{2}q_{1}} & { q_{1}^{2}-q_{2}^{2}-q_{3}^{2}+2q_{4}^{2}}
\end{array}\right)
\]
is the quaternion parametrization of a rotation matrix. This leads
to the following parametrization of the hand-eye calibration problem:

\vspace{0.2cm}\begin{problem}({\it qhec} method)\vspace{-0.3cm}

\[
\begin{array}{rl}
\textrm{minimize} & f_{2}(\mathbf{q}_{\X},\tx)=\\
 & \hspace{0.5cm}\sum_{i=1}^{n}\vnorm{\A_{i}\X(\mathbf{q}_{\X},\tx)-\X(\mathbf{q}_{\X},\tx)\B_{i}}^{2}\\
\textrm{subject to} & \mathbf{q}_{\X}^{\top}\mathbf{q}_{\X}=1,\\
 & q_{\X1}\geq0.
\end{array}
\]
\label{prob:qhec}\end{problem}The objective function $f_{2}$ of
Problem~\ref{prob:qhec} is a polynomial function of degree 4 and
it is composed of 85 monomials in 7 variables. Since the unit quaternions
are a double cover of $SO(3)$, function $f_{2}$ has at least two
global optima $f_{2}(\mathbf{q}_{\X}^{*},\tx^{*})=f_{2}(-\mathbf{q}_{\X}^{*},\tx^{*})$.
To help the SDP solver, we add the semialgebraic constraint $q_{\X1}\geq0$
to Problem~\ref{prob:qhec} to eliminate one of the global optima
in the majority of cases, that is when $q_{\X1}^{*}\neq0$. The algebraic
constraint $\mathbf{q}_{\X}^{\top}\mathbf{q}_{\X}=1$ enforces the
unity of the resulting quaternion.

\subsection{Dual Quaternion parametrization}

Dual quaternions are the algebraic counterparts of screws. They form
a Clifford algebra; a dual quaternion $\hat{\mathbf{q}}\in\mathbb{H}$
can be represented in the form
\[
\hat{\mathbf{q}}=\mathbf{q}+\epsilon\mathbf{q}',
\]
where $\mathbf{q},\mathbf{q}'\in\mathbb{Q}$ and $\epsilon$ is the
dual unit, $\epsilon\epsilon=0$, that commutes with every element
of the algebra. It is also convenient to write dual quaternions as
vectors $(\mathbf{q}^{\top},\mathbf{q}'^{\top})^{\top}$, since the
set of dual quaternions is equal to $\bR 8$. Addition of two dual
quaternions $\hat{\mathbf{p}},\hat{\mathbf{q}}\in\mathbb{H}$, $\hat{\mathbf{p}}=\left(\mathbf{p}^{\top},\mathbf{p}'^{\top}\right)^{\top}$,
$\hat{\mathbf{q}}=\left(\mathbf{q}^{\top},\mathbf{q}'^{\top}\right)^{\top}$
is equivalent to addition in $\bR 8$. Multiplication can be expressed
using quaternion multiplication as
\[
\hat{\mathbf{p}}\otimes\hat{\mathbf{q}}=((\mathbf{p}*\mathbf{q})^{\top},(\mathbf{p}*\mathbf{q}'+\mathbf{p}'*\mathbf{q})^{\top})^{\top}.
\]
Similar to the way rotations in $\bR 3$ can be represented by the
quaternions of unit length, rigid motions in $\bR 3$ can be represented
by unit dual quaternions~\cite{McCarthy90}: rotation represented
by a quaternion $\mathbf{p}\in\mathbb{Q}$ followed by translation
$\mathbf{t}\in\bR 3$ is represented by the dual quaternion
\[
\hat{\mathbf{q}}(\mathbf{p},\mathbf{t})=(\mathbf{p}^{\top},((0,\tfrac{1}{2}\mathbf{t}^{\top})^{\top}*\mathbf{p})^{\top})^{\top}.
\]
Unity of a dual quaternion $\hat{\mathbf{q}}$ can be expressed using
its conjugate $\hat{\mathbf{q}}^{*}=(\left.\mathbf{q}^{*}\right.^{\top},\left.\mathbf{q}'^{*}\right.^{\top})^{\top}$
as $\hat{\mathbf{q}}^{*}\otimes\hat{\mathbf{q}}=1$ or using the quaternion
parts as
\[
\mathbf{q}^{\top}\mathbf{q}=1\quad\textrm{and}\quad q_{1}q_{5}+q_{2}q_{6}+q_{3}q_{7}+q_{4}q_{8}=0.
\]

Let $\hat{\mathbf{a}}_{i}$, $\hat{\mathbf{b}}_{i}$ be the dual quaternion
representation of transformations $\A_{i}$, $\B_{i}$, respectively.
Since Equation~\ref{eq:hand-eye} can be expressed using dual quaternions
multiplications as
\begin{equation}
\hat{\mathbf{a}}_{i}\otimes\hat{\mathbf{q}}_{\X}=\hat{\mathbf{q}}_{\X}\otimes\hat{\mathbf{b}}_{i},\label{eq:hand-eye-dq}
\end{equation}
 where $\hat{\mathbf{q}}_{\X}$ is the dual quaternion representation
of the transformation $\X$, we can recover $\X$ using the following
formulation:

\vspace{0.2cm}\begin{problem}({\it dqhec} method)\vspace{-0.3cm}

\[
\begin{array}{rl}
\textrm{minimize} & f_{3}(\hat{\mathbf{q}}_{\X})=\sum_{i=1}^{n}\vnorm{\hat{\mathbf{a}}_{i}\otimes\hat{\mathbf{q}}_{\X}-\hat{\mathbf{q}}_{\X}\otimes\hat{\mathbf{b}}_{i}}^{2}\\
\textrm{subject to} & \mathbf{q}_{\X}^{\top}\mathbf{q}_{\X}=1,\\
 & q_{\X1}q_{\X5}+q_{\X2}q_{\X6}+q_{\X3}q_{\X7}+q_{\X4}q_{\X8}=0.\\
 & q_{\X1}\geq0.
\end{array}
\]
\label{prob:dqhec}\end{problem}

The objective function $f_{3}$ of Problem~\ref{prob:dqhec} is a
polynomial function of degree 4 and it is composed of 177 monomials
in 8 variables. Again, additional constraint $q_{\X1}\geq0$ is added
to eliminate one of the two global optima in most of the cases. Since
we are using dual quaternions directly, we have to take care of the
orientation ambiguity of the rotational quaternions $\mathbf{a}_{i},\mathbf{b}_{i}$
when converting matrices $\A_{i}$, $\B_{i}$ into dual quaternions
$\hat{\mathbf{a}}_{i}$, $\hat{\mathbf{b}}_{i}$. This is due to the
fact that even though quaternions $\mathbf{a}_{i},-\mathbf{a}_{i}$
and $\mathbf{b}_{i},-\mathbf{b}_{i}$ represent the same rotation,
the sign matters when Equation~\ref{eq:hand-eye} is expressed using
dual quaternions, \emph{i.e.},
\[
\hat{\mathbf{a}}_{i}(\mathbf{a}_{i},\mathbf{t}_{\A_{i}})\otimes\hat{\mathbf{q}}_{\X}-\hat{\mathbf{q}}_{\X}\otimes\hat{\mathbf{b}}_{i}(\mathbf{b}_{i},\mathbf{t}_{\B_{i}})\neq
\hat{\mathbf{a}}_{i}(-\mathbf{a}_{i},\mathbf{t}_{\A_{i}})\otimes\hat{\mathbf{q}}_{\X}-\hat{\mathbf{q}}_{\X}\otimes\hat{\mathbf{b}}_{i}(\mathbf{b}_{i},\mathbf{t}_{\B_{i}}).
\]
To check the ``compatibility'' of the quaternions $\mathbf{a}_{i}$
and $\mathbf{b}_{i}$, the \emph{screw congruence theorem}~\cite{Chen91}
can be used. It provides a necessary condition for the solution of
Equation~\ref{eq:hand-eye-dq}: 
\[
\hat{\mathbf{a}}_{i}\otimes\hat{\mathbf{q}}_{\X}=\hat{\mathbf{q}}_{\X}\otimes\hat{\mathbf{b}}_{i}\Rightarrow(a_{1},a'_{1})^{\top}=(b_{1},b'_{1})^{\top}.
\]
Dual quaternions were first applied to the hand-eye calibration problem
by Daniilidis and Bayro-Corrochano in~\cite{Danilidis96,Daniilidis98}.
Their method, however, does not make use of all of the information
in the camera and robot motions. Since the method assumes that $(a_{1},a'_{1})^{\top}=(b_{1},b'_{1})^{\top}$,
it only uses the imaginary parts $\bar{\mathbf{a}}_{i}$, $\bar{\mathbf{a}}'_{i}$,
$\bar{\mathbf{b}}_{i}$, $\bar{\mathbf{b}}'_{i}$. In~\cite{Malti10},
Malti and Barreto proposed a method based on dual quaternions that
uses also the real components, however their solution does not solve
for rotation and translation simultaneously.

\section{Simultaneous Hand-Eye and Robot-World Calibration}

Due to the apparent similarity of Equations~\ref{eq:hand-eye} and~\ref{eq:robot_world},
analogies to Problems~\ref{prob:uvhec}, \ref{prob:qhec}, and \ref{prob:dqhec}
can be formulated for the simultaneous hand-eye and robot-world calibration
problem.

\subsection{Orthonormal parametrization}

Problem~\ref{prob:uvhewbc} is based on orthonormal parametrization
of Equation~\ref{eq:robot_world}. The objective function $f_{4}$
is a polynomial function of degree 4 and it is composed of 280 monomials
in 18 variables.

\vspace{0.2cm}\begin{problem}({\it uvherwc} method)\vspace{-0.3cm}

\[
\begin{array}{rl}
\textrm{minimize} & f_{4}(\mathbf{u}_{\X},\mathbf{v}_{\X},\tx,\mathbf{u}_{\Z},\mathbf{v}_{\Z},\tz)=\\
 & \hspace{0.5cm}\sum_{i=1}^{m}\vnorm{\A_{i}'\X(\mathbf{u}_{\X},\mathbf{v}_{\X},\tx)-\Z(\mathbf{u}_{\Z},\mathbf{v}_{\Z},\tz)\B_{i}'}^{2}\\
\textrm{subject to} & \mathbf{u}_{\X}^{\top}\mathbf{u}_{\X}=1,\mathbf{v}_{\X}^{\top}\mathbf{v}_{\X}=1,\mathbf{u}_{\X}^{\top}\mathbf{v}_{\X}=0,\\
 & \mathbf{u}_{\Z}^{\top}\mathbf{u}_{\Z}=1,\mathbf{v}_{\Z}^{\top}\mathbf{v}_{\Z}=1,\mathbf{u}_{\Z}^{\top}\mathbf{v}_{\Z}=0.
\end{array}
\]
\label{prob:uvhewbc}\end{problem}

\subsection{Quaternion parametrization}

Problem~\ref{prob:qhewbc} is based on quaternion parametrization
of Equation~\ref{eq:robot_world}. The objective function $f_{5}$
is a polynomial function of degree 4 and is composed of 209 monomials
in 14 variables.

\vspace{0.2cm}\begin{problem}({\it qherwc} method)\vspace{-0.3cm}

\[
\begin{array}{rl}
\textrm{minimize} & f_{5}(\mathbf{q}_{\X},\tx,\mathbf{q}_{\Z},\tz)=\\
 & \hspace{0.5cm}\sum_{i=1}^{m}\vnorm{\A_{i}'\X(\mathbf{q}_{\X},\tx)-\Z(\mathbf{q}_{\Z},\tz)\B_{i}'}^{2}\\
\textrm{subject to} & \mathbf{q}_{\X}^{\top}\mathbf{q}_{\X}=1,\, q_{\X1}\geq0,\\
 & \mathbf{q}_{\Z}^{\top}\mathbf{q}_{\Z}=1,\, q_{\Z1}\geq0.
\end{array}
\]
\label{prob:qhewbc}\end{problem}

\subsection{Dual Quaternion parametrization}

Problem~\ref{prob:dqhewbc} is based on quaternion parametrization
of Equation~\ref{eq:robot_world}. The objective function $f_{6}$
is a polynomial function of degree 2 and it is composed of 112 monomials
in 16 variables.

\vspace{0.2cm}\begin{problem}({\it dqherwc} method)\vspace{-0.3cm}

\[
\begin{array}{rl}
\textrm{minimize} & f_{6}(\hat{\mathbf{q}}_{\X})=\sum_{i=1}^{m}\vnorm{\hat{\mathbf{a}}_{i}'\otimes\hat{\mathbf{q}}_{\X}-\hat{\mathbf{q}}_{\Z}\otimes\hat{\mathbf{b}}_{i}'}^{2}\\
\textrm{subject to} & \mathbf{q}_{\X}^{\top}\mathbf{q}_{\X}=1,\, q_{\X1}\geq0,\mathbf{q}_{\Z}^{\top}\mathbf{q}_{\Z}=1,\, q_{\Z1}\geq0,\\
 & q_{\X1}q_{\X5}+q_{\X2}q_{\X6}+q_{\X3}q_{\X7}+q_{\X4}q_{\X8}=0,\\
 & q_{\Z1}q_{\Z5}+q_{\Z2}q_{\Z6}+q_{\Z3}q_{\Z7}+q_{\Z4}q_{\Z8}=0.
\end{array}
\]
\label{prob:dqhewbc}\end{problem}As is the case of Problem~\ref{prob:dqhec},
we need to take care of the orientation ambiguity of the rotational
quaternions $\mathbf{a}_{i},\mathbf{b}_{i}$. This time however, the
screw congruence theorem does not apply. The obvious solution it to
try all of the $2^{m}$ sign combinations and keep the combination
with the smallest value of $f_{6}(\hat{\mathbf{q}}_{\X}^{*})$. Running
the SDP optimization $2^{m}$ times can be quite computationally expensive,
so in our experiments we use the dual quaternion method of Li~\cite{Li10}---which
needs to be run $2^{m}$ times in practice as well, since it suffers
from the same quaternion ambiguity problem---to recover the sign combination
before running the SDP solver, because it runs faster. The sign problem
is also inherent to the method of Dornaika~\cite{Dornaika98}.

\section{Experiments}

In this section we present both synthetic and real data experiments
to validate the proposed calibration methods. For the real data experiment
we used a Motoman MA1400 serial 6-DOF manipulator with an Asus Xtion
Pro sensor rigidly attached to it. The Xtion Pro sensor was equipped
with a camera with the resolution of $\textrm{640}\times\textrm{480}$
pixels. We simulated the same setup in the synthetic experiment in
order to better judge the result of the experiment with real data.

\subsection{Implementation}

\begin{figure*}
\begin{centering}
\subfloat[]{

\centering{}\includegraphics{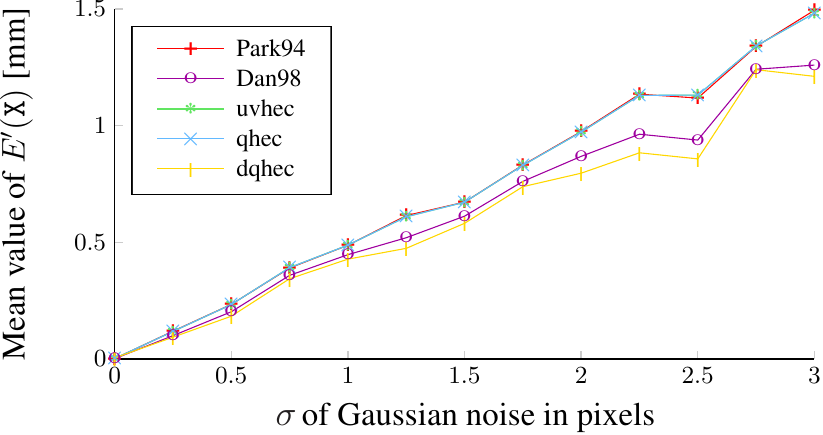}}\subfloat[]{

\centering{}\includegraphics{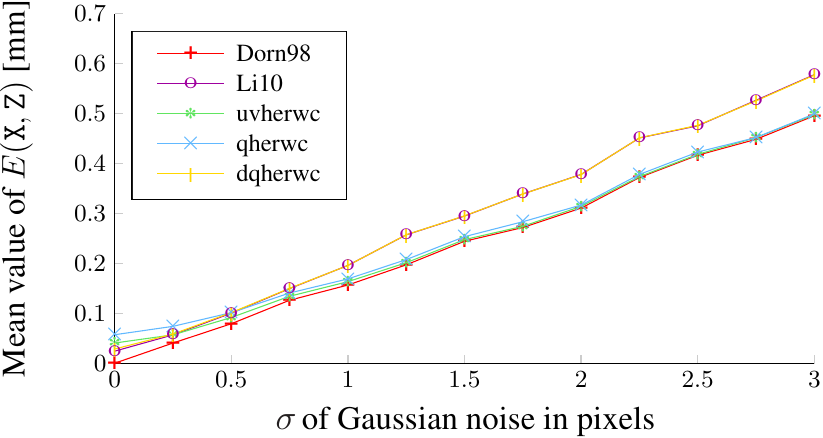}}\vspace{-0.8cm}
\\
\subfloat[]{

\centering{}\includegraphics{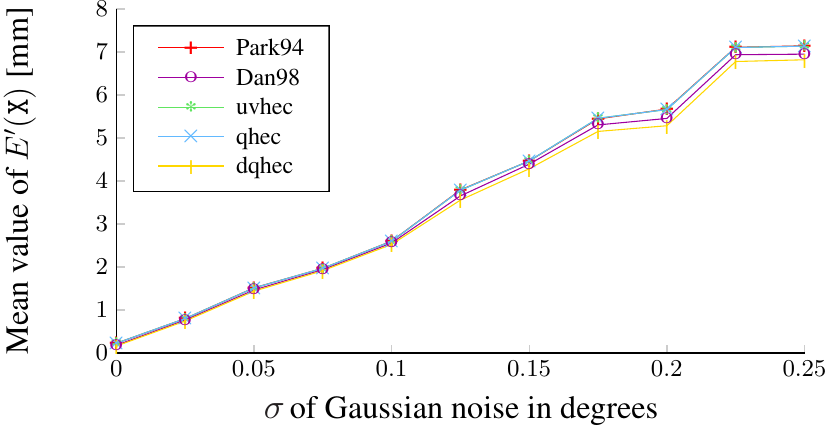}}\subfloat[]{

\centering{}\includegraphics{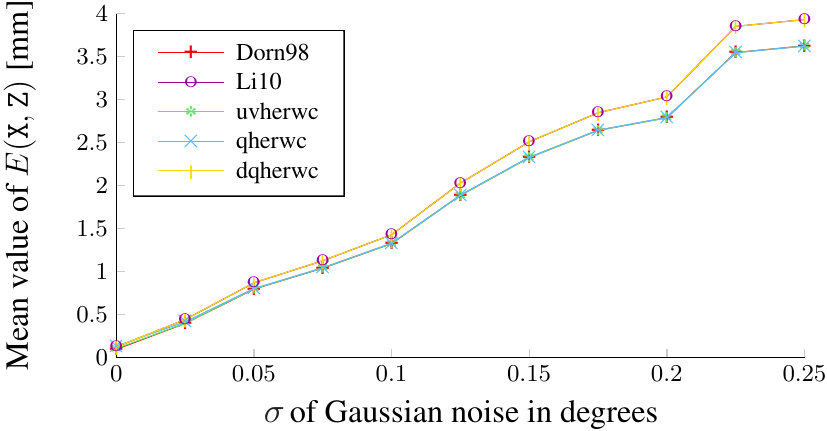}}
\par\end{centering}

\centering{}\caption{\emph{Synthetic data experiment results}. (a--b) Image noise experiment.
(c--d) Joint noise experiment.\label{fig:synth-exp}}
\end{figure*}

We implemented all 6 calibration methods in \textsc{Matlab} using
GloptiPoly~\cite{Henrion09}, an interface that automatically constructs
LMI relaxations of polynomial problems and converts them into a format
understandable by the SDP solver SeDuMi~\cite{Sturm99}. Further,
GloptiPoly can also recover the solution to the original polynomial
problem and certify its optimality. Using YALMIP toolbox~\cite{yalmip04},
the SeDuMi format can be further converted into the input formats
of several other SDP solvers. In our experiments, we used the LMI
relaxations of the second order and SeDuMi 1.3 and MOSEK~\cite{mosek}
7.0 as SDP solvers. GloptiPoly certified that global optimum has been
reached by all the methods for all the problem instances encountered
during the experiments. The following table sums up the relevant information
on the proposed methods:

\begin{center}
\begin{tabular}{|r|r|r|r|r|r|r|}
\hline 
\begin{sideways}
Method~
\end{sideways} & \begin{sideways}
Variables~
\end{sideways} & \begin{sideways}
Degree
\end{sideways} & \begin{sideways}
Monomials~
\end{sideways} & \begin{sideways}
Moments~
\end{sideways} & \begin{sideways}
Time$^{1}$(s)
\end{sideways} & \begin{sideways}
Time$^{2}$(s)~
\end{sideways}\tabularnewline
\hline 
\hline 
\emph{uvhec} & 9 & 4 & 124 & 715 & 1.45 & 0.45\tabularnewline
\hline 
\emph{qhec} & 7 & 4 & 85 & 330 & 0.48 & 0.29\tabularnewline
\hline 
\emph{dqhec} & 8 & 4 & 177 & 495 & 0.99 & 0.46\tabularnewline
\hline 
\emph{uvherwc} & 18 & 4 & 280 & 7315 & 953.76 & 60.89\tabularnewline
\hline 
\emph{qherwc} & 14 & 4 & 209 & 3060 & 68.38 & 7.91\tabularnewline
\hline 
\emph{dqherwc} & 16 & 2 & 112 & 4845 & 309.25 & 16.85\tabularnewline
\hline 
\multicolumn{7}{r}{{\scriptsize $^{{\scriptscriptstyle 1}}$SeDuMi,$^{{\scriptscriptstyle 2}}$MOSEK }}\tabularnewline
\end{tabular}
\par\end{center}

\noindent{}To obtain the timings, we used a 3.5GHz Intel Core i7
based desktop computer running 64-bit Linux. The presented times include
time spent in the SDP solver as well as GloptiPoly and YALMIP overheads.
To gauge the sizes of the SDP problems involved, the column ``Moments''
specifies the number of moments (dimension of vector $\mathbf{y}$
in Problem~\ref{prob:mpoly_sdp}) in the second order LMI relaxation
for the respective method. Note, that even though the objective functions
$f_{i}$ depend on $n$ or $m$, \emph{i.e.}, on the number of relative
motions or poses, the number of monomials does not. This means that
the sizes of the LMI relaxations and the sizes of the resulting SDP
problems also do not depend on $n$ or $m$ and are constant.

As a pre-step to all of the methods, we scale the translations $\tai i,\tbi i$
by the factor of $\alpha=\max_{i}\left\{ \vnorm{\tai i},\vnorm{\tbi i}\right\} $,
so that the length of the longest translation is 1. This helps with
the convergence of the SDP solver and removes the influence of the
chosen physical units on the accuracy of the result. In cases where
$\tx$ and $\tz$ are explicitly optimized, we add one or two more
constraints $\tx^{\top}\tx\leq2$, $\tz^{\top}\tz\leq10$. These constraints
result from the way our experiments were constructed and are not technically
necessary, however, they help the SDP solvers to further speed-up
the converge. In the experiments, we set the SeDuMi parameter $\texttt{eps}=10^{-20}$
and the MOSEK parameters $\texttt{MSK\_DPAR\_INTPNT\_CO\_TOL\_\{P}\!\texttt{|}\!\texttt{D\}FEAS}\!=\!10^{-20}$
.

The source code as well as examples are available at

\url{\small http://cmp.felk.cvut.cz/{\raise.17ex\hbox{$\scriptstyle\mathtt{\sim}$}}hellej1/mpherwc/}

\subsection{Synthetic Experiments}

\noindent{}In the synthetic experiments, we investigated the influence
of both image and joint noises on the calibration accuracy. We placed
a virtual planar calibration target consisting of a grid of $\textrm{16}\!\times\!\textrm{16}$
known points in front of a simulated MA1400 serial manipulator. The
distance between the points was set to 12.5~mm in each direction,
making the calibration target $\textrm{200}\!\times\!\textrm{200}\,\textrm{mm}$
in size. 

To create a calibration task, we started with 9 camera poses $\mathtt{A}_{i}'$
randomly generated onto half-sphere of radius of $\sim\!\textrm{30}\,\textrm{mm}$
oriented as to face approximately the center of the calibration target.
To simulate the typical situation when the camera faces approximately
the same direction as the robot's end effector, a hand-eye rotation
$\Rx$ was randomly generated so that the rotation corresponded to
identity to a up to 5$^{\circ}$ degrees difference in each axis.
The translation $\tx$ was randomly generated to move the camera up
to $\sim\!\textrm{200}\,\textrm{mm}$ away from end effector. The
arm poses $\mathtt{B}_{i}'$ were computed using the camera poses
and the hand-eye transformation as $\mathtt{X}^{-1}\mathtt{A}_{i}'$.
Finally, we generated a robot-world transformation $\mathtt{Z}$ using
a random rotation and a random translation up to $\sim\!\textrm{2000}\,\textrm{mm}$
in length and used $\Z$ to transform the grid points from the robot
into the world coordinate system. We generated 10 sets of 9 camera
positions, 10 hand-eye transformations and 1 robot-world transformation
and combined them into 100 calibration tasks. For every calibration
task, we also computed all possible relative movements $\A_{i}$,
$\B_{i}$, $i=\textrm{1},\dots,\textrm{36}$ to be used by the hand-eye
calibration methods. In order to better judge the accuracy the proposed
methods, we implemented several hand-eye and robot-world calibration
methods to compare them against: \emph{Park94} \cite{Park94} and
\emph{Dan98} \cite{Daniilidis98}, as representatives of hand-eye
calibration methods, and \emph{Dorn98} \cite{Dornaika98} and \emph{Li10},
the dual quaternion variant of \cite{Li10}, as representatives of
simultaneous hand-eye and robot-world calibration methods.

\begin{figure}
\centering{}\subfloat[]{

\centering{}\includegraphics[scale=0.41]{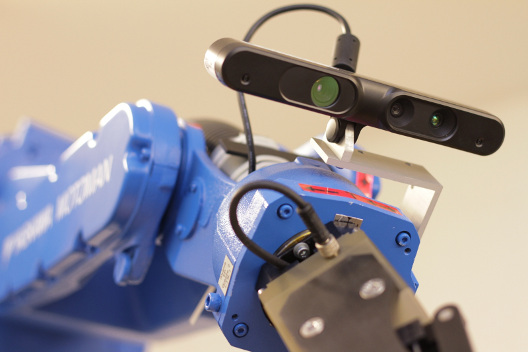}}\subfloat[]{\centering{}\includegraphics[scale=0.3]{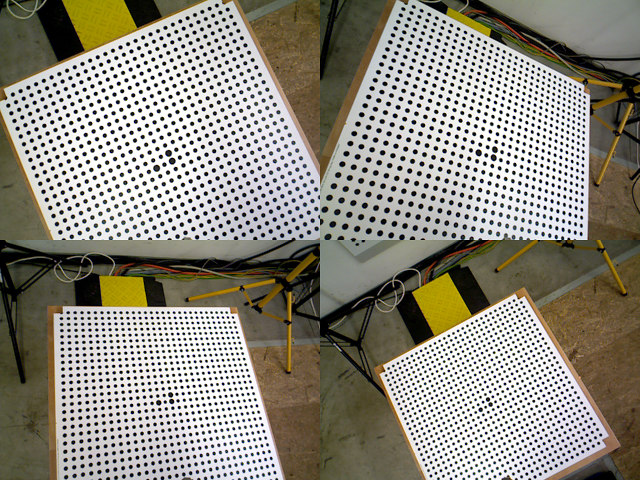}}\caption{\emph{Real data experiment setup}. (a) Detail of the Xtion Pro sensor
and the robot's end effector. (b)~Examples of input images.\label{fig:real-exp}}
\end{figure}

\subsubsection*{\textmd{Image Noise Experiment}}

To simulate the influence of image noise, we projected the calibration
points into $\textrm{640}\times\textrm{480}$ pixel images using the
generated camera poses $\A_{i}'$ and corrupted the image projections
$\mathbf{u}_{ij}$ in each calibration task with Gaussian noise in
13~noise levels, $\sigma\in\langle\textrm{0},\textrm{3}\rangle$
px in $\nicefrac{\textrm{1}}{\textrm{4}}\,\textrm{px}$ steps. Finally,
we recovered noised camera poses $\A_{i}'$ using EPnP algorithm~\cite{Lepetit09}. 

In order to evaluate the performance of the hand-eye and robot-world
calibration methods \emph{uvherwc}, \emph{qherwc}, and \emph{dqherwc},
we have sampled the virtual workspace of the robot---a cube of size
$\textrm{70}\!\times\!\textrm{70}\!\times\!\textrm{70}$~cm around
the calibration target---with uniformly distributed $\ell=\textrm{9240}$
points $\mathbf{Y}_{j}$. Using these points, measured in the robot
coordinate frame, we defined a calibration error measure $E(\X,\Z)$
that expresses the error of the calibration as the difference of the
positions the workspace points transformed into the camera coordinate
frame using the computed transformations $\X$, $\Z$ and the ground
truth transformations $\X^{\mathrm{gt}}$, $\Z^{\mathrm{gt}}$ . Formally,
\[
E(\X,\Z)={\textstyle \frac{1}{9\ell}\sum_{i=1}^{9}\sum_{j=1}^{\ell}}\left\Vert \mathtt{X}\mathtt{B}_{i}^{'-1}\Z^{-1}\Z^{\mathrm{gt}}\mathbf{Y}_{j}-\mathtt{X}^{\mathrm{gt}}\mathtt{B}_{i}^{'-1}\mathbf{Y}_{j}\right\Vert .
\]
To evaluate the performance of the hand-eye calibration methods \emph{uvhec},
\emph{qhec}, and \emph{dqhec}, we do not have to transform the points
$\mathbf{Y}_{j}$ into the world coordinate frame first, so we defined
a slightly simpler 3D error function
\[
E'(\X)={\textstyle \frac{1}{36\ell}\sum_{i=1}^{36}\sum_{j=1}^{\ell}}\left\Vert \mathtt{X}\mathtt{B}_{i}^{'-1}\mathbf{Y}_{j}-\mathtt{X}^{\mathrm{gt}}\mathtt{B}_{i}^{'-1}\mathbf{Y}_{j}\right\Vert .
\]

Figure~\ref{fig:synth-exp}(a) shows the mean of the values $E'(\X)$
for all 100 calibration tasks for different methods and image noise
levels. It shows the best performance by the method \emph{dqhec} followed
by \emph{Dan98} and  finally by \emph{Park94}, \emph{uvhec}, and \emph{qhec},
all with the same performance level. In this experiment, we used SeDuMi
as the SDP solver.

Figure~\ref{fig:synth-exp}(b) shows the same statistics for the
values of $E(\X,\Z)$ and suggest quite comparable performance of
methods \emph{Dorn98}, \emph{uvhewrc}, and \emph{qhewbc}. This time,
the dual quaternion formulations \emph{Li10} and \emph{dqherwc} performed
worse that the other methods. In this experiment, we used MOSEK as
the SDP solver. This resulted in much faster convergence times, but
also in slightly worse accuracy for the lower noise levels.

\begin{figure}
\begin{centering}
\subfloat[]{

\centering{}\includegraphics[scale=1]{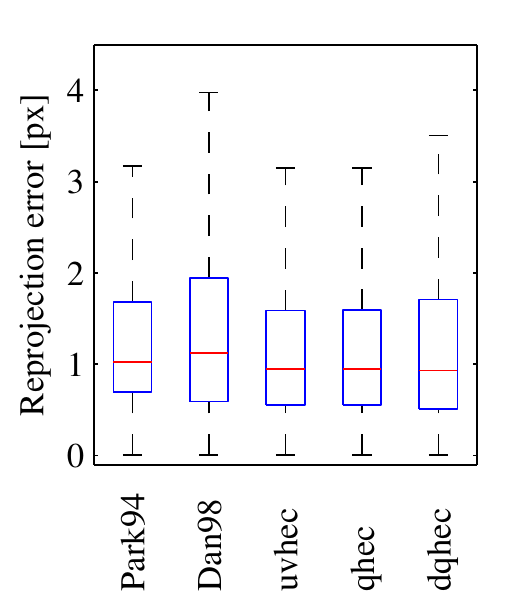}}\hspace{0.5cm}\subfloat[]{\centering{}\includegraphics[scale=1]{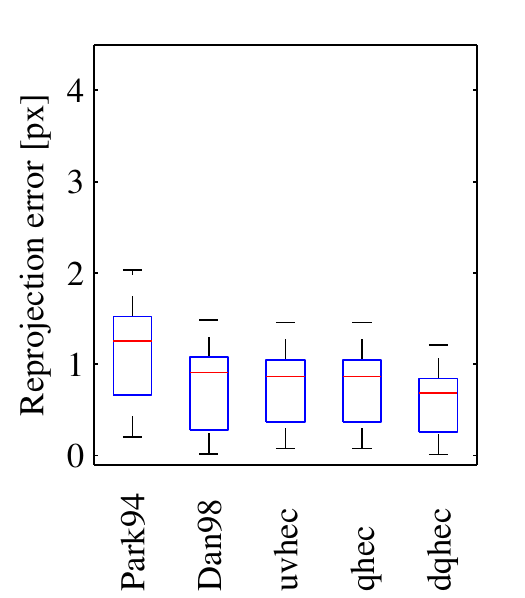}}\vspace{-0.3cm}
\\
\subfloat[]{

\centering{}\includegraphics[scale=1]{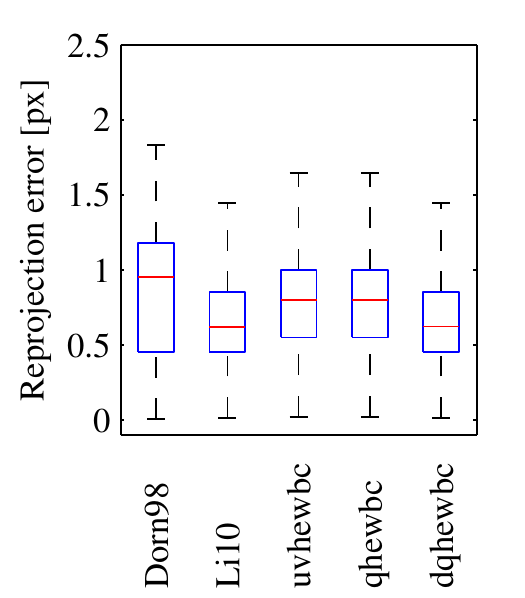}}\hspace{0.5cm}\subfloat[]{\centering{}\includegraphics[scale=1]{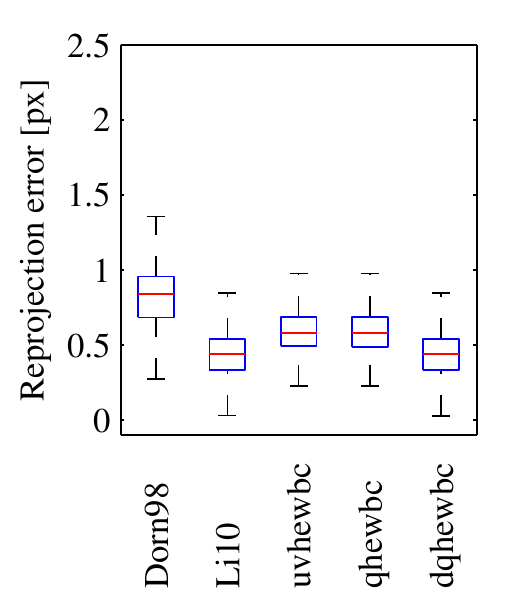}}
\par\end{centering}

\centering{}\caption{\emph{Real data experiment resuts}. (a) Reprojection errors $e_{ij}'$
of the calibration image set (b) Reprojection errors $e_{ij}'$ of
the validation image set. (a) Reprojection errors $e_{ij}$ of the
calibration image set (b) Reprojection errors $e_{ij}$ of the validation
image set. (the red line marks the median, the edges of the box are
the 25th and 75th percentiles) \label{fig:real-exp-res}}
\end{figure}

\subsubsection*{\textmd{Joint Noise Experiment}}

In this experiment, we simulated the performance in the presence of
joint noise. We started with the same 100 calibration tasks as in
the case of the image noise experiment. Further, we recovered the
joint coordinates~\cite{Siciliano08} of the virtual MA1400 manipulator
for every pose $\mathtt{B}_{i}'$ with respect to the Denavit\textendash{}Hartenberg
convention. Next, we corrupted the joint coordinates by random offsets---we
used the same offsets for the joint coordinates of the same task---in
11 noise levels, $\sigma\in\langle\textrm{0},\textrm{0.25}\rangle\,\textrm{deg}$
in steps of 0.025 degrees. Finally, we recovered noised poses $\mathtt{B}_{i}'$
using the forward kinematics. To further simulate the real world conditions,
we corrupted projections $\mathbf{u}_{ij}$ by image noise of $\sigma=\textrm{0.5}\,\textrm{px}$
for every joint noise level. We evaluated the performance using the
same error measures $E(\X,\Z)$ and $E'(\X)$.

Figures~\ref{fig:synth-exp}(c) and \ref{fig:synth-exp}(d) show
the mean of the values $E'(\X)$ and $E'(\X,\Z)$. Again, the method
\emph{qhec} outperforms its competitors, whereas the dual quaternion
formulations lose on the rest in the simultaneous hand-eye and robot-world
calibration experiment.

\subsection{Real Data Experiment}

For this experiment, we used a real MA1400 serial manipulator with
a Xtion Pro sensor attached to its 5th link, see Figure~\ref{fig:real-exp}(a).
We manipulated the robotic arm into 10 poses $\B_{i}'$ and acquired
a calibration image set consisting of the same number of images of
a $\textrm{30}\!\times\!\textrm{30}$ points calibration grid. The
grid was placed $\sim\!\textrm{1}\,\textrm{m}$ in front of and $\sim\!\textrm{0.5}\,\textrm{m}$
above the base of the robotic arm. The point spacing of the calibration
grid was $\textrm{24}\,\textrm{mm}$. We also acquired a validation
image set consisting of 3 images. See Figure~\ref{fig:real-exp}(b)
for example images from the calibration sequences. Next, we used OpenCV~\cite{OpenCV}
library to obtain the internal calibration matrix $\mathtt{K}$~\cite{Hartley04}
of the sensor as well as the camera poses $\A_{i}'$, $i=\textrm{1},\dots,\textrm{13}$.

Since there was no ground truth information available, we had to use
a performance measure different from $E(\X,\Z)$ and $E'(\X)$ used
in the synthetic data experiments. Suppose, that a~function $\mathcal{P}(\mathbf{Y},\A'_{i},\mathtt{K})$
projects calibration grid points $\mathbf{Y}_{j}\in\mathbb{R}^{3}$,
$j=1,\dots,\textrm{900}$ in the world coordinate frame into the $i$-th
image taken by a camera described by its pose $\A'_{i}$ and internal
camera calibration matrix $\mathtt{K}$. The reprojection error of
the point $\mathbf{Y}_{j}$ is defined as $e_{ij}(\mathbf{u}_{i},\mathbf{Y}_{j},\A'_{i},\mathtt{K})=\vnorm{\mathbf{u}_{ij}-\mathcal{P}(\mathbf{Y}_{j},\A'_{i},\mathtt{K})}$,
where $\mathbf{u}_{ij}$ are the pixel coordinates of the point $\mathbf{Y}_{j}$
in the $i$-th image. In case a calibration method recovers both $\X$
and $\Z$, we can judge the quality of the calibration by expressing
the reprojection error as $e_{ij}(\mathbf{u}_{i},\mathbf{Y}_{j},\Z\B_{i}\X^{-1},\mathtt{K})$.
However, in case only transformation $\X$ is recovered, the camera
pose $\A'_{i}$ can be expressed only with the help of an additional
pose $\A'_{i}=\A'_{k}\X\B_{k}^{\prime-1}\B'_{i}\X^{-1}$ and we have
to define modified reprojection error $e'_{ij}=\frac{1}{m-1}\sum_{k=1,k\neq i}^{m}e_{ij}(\mathbf{u}_{ij},\mathbf{Y}_{j},\A'_{k}\X\B_{k}^{\prime-1}\B'_{i}\X^{-1},\mathtt{K})$.

Figure~\ref{fig:real-exp-res}(a) shows the statistics of the modified
reprojection errors $e_{ij}^{'}$ computed from the calibration image
set created by the \textsc{Matlab} function \texttt{\small boxplot}
for methods \emph{Park94}, \emph{Dan99}, \emph{uvhec}, \emph{qhec},
and \emph{dqhec}. Figure~\ref{fig:real-exp-res}(b) shows the same
statistics, this time for the validation image set. As is the case
of synthetics data experiments, \emph{dqhec} slightly outperforms
its competitors. Figures~\ref{fig:real-exp-res}(c,d) show the reprojection
errors $e_{ij}$ computed for the methods \emph{Dorn98}, \emph{Li10},
\emph{uvherwc}, \emph{qherwc}, and \emph{dqherwc} for the calibration
and validation image sets, respectively.

\section{Conclusion}

In this paper, we showed that the method of convex LMI relaxations
can be naturally applied to the hand-eye and robot-world calibration
problems. We presented three hand-eye and three hand-eye and robot-world
calibration parametrizations and by applying the method of convex
LMI relaxations we obtained globally optimal solutions. These formulations
provide a new insight into the behavior and complexity of the original
problems. The \emph{qhec} parametrization showed the best performance
overall. Methods \emph{uvherwc} and \emph{qherwc} do not necessarily
provide more accurate results than the previously proposed methods,
however, since they do not suffer from the quaternion sign ambiguity,
the running time is constant and not exponential in the number of
poses.

\section*{Acknowledgements}

The authors were supported by the EC under projects FP7-ICT-288553 CloPeMa, FP7-SPACE-2012-312377 PRoViDE, and by The Technology Agency of the Czech Republic under the project TA03010398 RoMeSy. This work also benefited from discussions with Florian Bugarin.

\end{document}